\documentclass[11pt]{article}

\usepackage[final]{acl}

\usepackage{times}
\usepackage{latexsym}
\usepackage{amsmath}
\usepackage{algorithm}
\usepackage{algpseudocode}
\usepackage{booktabs}
\usepackage{multirow}
\usepackage{multicol}
\usepackage{pgfplots}
\usepackage{tikz}
\usepackage{amsthm} 
\usepackage{amsfonts} 
\usepackage{amssymb}
\usepackage{multirow}
\usepackage{dcolumn}
\usepackage{array}
\usepackage{graphicx}
\usepackage{titlesec}
\usepackage{enumitem}
\usepackage{url}
\usepackage{authblk}
\usepackage{caption}
\usepackage{booktabs}
\usepackage{verbatim}
\usepackage{tcolorbox}
\usepackage{xcolor}
\usepackage{soul}
\pgfplotsset{compat=1.18}
\usepgfplotslibrary{units}
\usetikzlibrary{arrows.meta, positioning, fit, backgrounds}
\usepackage{color,hyperref}
\definecolor{darkblue}{rgb}{0.0,0.0,0.3}
\hypersetup{colorlinks,breaklinks,
linkcolor=darkblue,urlcolor=darkblue,
anchorcolor=darkblue,citecolor=darkblue}
\usepackage{url}
\usepackage{graphicx}

\usepackage[noabbrev,nameinlink]{cleveref}

\usepackage[T1]{fontenc}
\usepackage[utf8]{inputenc}

\usepackage{microtype}

\usepackage{inconsolata}

\usepackage{graphicx}

\title{A benchmark for end-to-end zero-shot biomedical relation extraction \\ with LLMs: experiments with OpenAI models}

\author{
 \textbf{Aviv Brokman},
 \textbf{Xuguang Ai},
  \textbf{Yuhang Jiang},
  \textbf{Shashank Gupta},
  \textbf{Ramakanth Kavuluru}
\\
\\
Division of Biomedical Informatics, Department of Internal Medicine\\
University of Kentucky, USA.
\\
 \small{
    \textit{Correspondence:} \texttt{ramakanth.kavuluru@uky.edu}
}}
\pgfplotsset{compat=1.18}
\begin{document}
\maketitle
\begin{abstract}

Extracting relations from scientific literature is a fundamental task in biomedical NLP because entities and relations among them drive hypothesis generation and knowledge discovery. As literature grows rapidly, relation extraction (RE) is indispensable to curate knowledge graphs to be used as computable structured and symbolic representations. With the rise of LLMs, it is pertinent to examine if it is better to skip tailoring supervised RE methods, save annotation burden, and just use zero shot RE (ZSRE) via LLM API calls. In this paper, we propose a benchmark with seven biomedical RE datasets with interesting characteristics and evaluate three Open AI models (GPT-4, o1, and GPT-OSS-120B) for end-to-end ZSRE. We show that LLM-based ZSRE is inching closer to supervised methods in performances on some datasets but still struggles on complex inputs expressing multiple relations with different predicates. Our error analysis reveals scope for improvements. 
\end{abstract}

\section{Introduction}

Do we need training data to perform relation extraction (RE)? Since ChatGPT was introduced in December 2022, this has been a prominent question on minds of many NLP researchers, especially those that focus on structured information extraction from biomedical literature. With the recent success of zero-shot methods in other areas of NLP, RE is ripe for investigation, and biomedicine is a particularly compelling domain as relations are central to knowledge discovery. 

% What is RE?
RE is the high-value NLP task of identifying semantic relationships between entities within text. Consider the following sentence taken from the drug combination extraction (DCE) \cite{tiktinsky-etal-2022-dataset} dataset: \textit{``Furthermore, in non-metastatic castration-resistant prostate cancer (M0 CRPC), two second-generation anti-androgens, apalutamide, and enzalutamide, when used in combination with ADT, have demonstrated a significant benefit in metastasis-free survival.''}
Two beneficial drug combinations are described here: (1) \texttt{apalutamide} with \texttt{ADT}, and (2) \texttt{enzalutamide} with \texttt{ADT}. In RE, we want to extract these relations into a structured form; in this case a tuple of drugs administered in combination along with a signifier of the normative effect of the drug: 

\noindent\parbox{\linewidth}{
\begin{itemize}
    \item \texttt{\{drugs: (apalutamide, ADT), effect: positive\}}
    \item \texttt{\{drugs: (enzalutamide, ADT), effect: positive\}}
\end{itemize}
} 

\noindent Thus, RE can be viewed as the conversion of unstructured data into structured data representing relationships between entities. 
In biomedicine, entities of interest are mainly genes, mutations, proteins, chemicals, drugs, diseases, and symptoms. The relationships that can hold between them are myriad, but some obvious relationships of importance are drug interactions, protein interactions, disease-causing mutations and chemicals, drug side effects, and disease-treating drugs. 
%Typical texts used for mining biomedical relations are journal abstracts and clinical notes. 
Many RE efforts assume that the entities and their types are already provided as part of the input. They do RE by giving the input text and a pair of entity spans in it and ask for the relation type linking them. The technical name for this would be relation classification (RC).  However, for RE to be fully automated and evaluated fairly, the input must be just free text and the burden of (a) spotting the entities and (b) linking them with predicates, both fall on the method. This is called end-to-end RE (E2E RE) and is much harder than RE when entities are pre-annotated. For the rest of this paper, whenever we refer to RE we mean the end-to-end variety. 

% importance of RE
Biomedical publications  are being generated at breakneck speed --- PubMed indexes nearly 40 million articles and over four thousand more are indexed daily. Biomedical relations are so valuable that there are teams of workers employed to read biomedical text and populate databases with them. With so much text available, automating RE would allow us to mine biomedical relations at scale, rapidly enlarging databases. 

\subsection{Zero-shot RE}
Traditionally, RE has been conducted by fine-tuning with hundreds to thousands of examples. In this paradigm, every narrow RE task requires a dataset to be curated for it. Dataset curation for biomedical RE is a laborious process --- annotators need highly specialized knowledge, it takes time to develop clear and consistent annotation guidelines, and annotators need to be trained on the guidelines \cite{luo2022biored, li2016biocreative}. The process is so laborious for annotators that, typically, named entity recognition and RE tools are used to make suggestions to annotators to speed up the process \cite{luo2022biored, li2016biocreative}. For all these reasons, few-shot (FS) and zero-shot (ZS) RE are desirable. They remove the need for a training dataset, which is usually the largest in a training/validation/test split. If modeling choices do not need to be made and if performance need not be measured, validation and test sets could be omitted as well.
The promise of high quality low-resource RE is a proliferation of databases and an increase in their richness and reliability.  

Generative large language models (LLMs) have grown to dominate NLP research activity in recent years, with new multi-billion parameter models being released regularly. Owing to large amounts of diverse training data, vast quantities of parameters, and techniques that align models with human goals like instruction finetuning \cite{instruct-gpt, wei-ift, sanh-ift} and RLHF \cite{ouyang2022training}, these models can now perform a wide array of tasks, in a ZS manner, with impressive results.  
However, LLMs are not adept at producing long output in a consistent format, a key challenge when converting generated text into structured relations, without specific guiding mechanisms. To address this, previous studies have employed two main strategies: (1) prompting the language model to generate text in a predetermined format, followed by the use of predefined regular expressions to extract relation components \cite{luo2022biogpt, gupta2025comparison}, or (2) directly specifying a structured output within the prompt itself \cite{wadhwa-etal-2023-revisiting}. Though these approaches have shown promise,  \citet{wadhwa-etal-2023-revisiting} found that even for few-shot sentence-level RE,  GPT-3.5 often generates plausible relations that, while recognizable to humans as correct, do not precisely match the gold standard relations. This discrepancy should be expected to be more pronounced in document-level ZS settings. Without fine-tuning, fulfilling synergistically demanding requirements of long, exact, consistent extraction from text becomes more challenging. 

Zero-shot relation extraction (ZSRE) poses a unique challenge to LLMs because it requires the generation of long, exact text in a consistent format. This makes it much more challenging to generate an exact output than most other exact-output tasks NLP researchers have been tackling, such as question answering (QA), where the answers are a single token or a phrase~\cite{touvron2023llama, wang-etal-2023-self-instruct, openai2023gpt4}. Also, ZS generation has been successful at tackling problems where long text must be generated, like summarizing \cite{touvron2023llama, wang-etal-2023-self-instruct, openai2023gpt4}. In such tasks, there is no single correct generation, so the fact that LLMs produce diverse output is not a problem; these are often evaluated by humans or judgment by other more  LLMs (though this practice is controversial). But in the real-world scenario of RE from a document, there may be many relations present (the BioRED \cite{luo2022biored} dataset contains abstracts with $> 50$ relations), and the LM must generate long text with stringent requirements on what the text must consist of.

Due to the aforementioned challenges, LLMs generate relations that are correct to a human but do not match annotations in the test dataset exactly, which artificially deflates calculated performance.  For example, if an annotated relation contains the entity \texttt{hypertensive} as the disease in a relation and an LM extracts \texttt{hypertension}, this would be considered incorrect in the usual performance evaluation.  \citet{wadhwa-etal-2023-revisiting} deal with this problem using manual evaluation of their relation extraction systems, but this is not desirable because (1) it is expensive and (2) because we want a system that can be used at scale to populate databases automatically, without human intervention. 

% What we do

\subsection{Related work}
Most RE methods focus on constructing embeddings for candidate relations, followed by a classification step. A parallel line of research has developed around the use of copy-mechanisms in a sequence-to-sequence (seq2seq) framework. Seq2seq tasks involve generating an output/target sequence from a given input/source sequence. This method is predominantly favored in areas like machine translation, where the format aligns naturally with the task. However, researchers have adapted RE to fit into the seq2seq paradigm by transforming structured relations into predefined sequences of tokens \cite{zeng-etal-2018-extracting, zhang-etal-2020-minimize, Nayak_Ng_2020, Zeng_Zhang_Liu_2020, giorgi-etal-2022-sequence}. For instance, \citet{giorgi-etal-2022-sequence} transform the relation \texttt{\{gene: ESR1, disease: schizophrenia, predicate:association\}} into the sequence \texttt{ESR1 @gene schizophrenia @disease @association}\footnote{This is a slight adaptation from the original paper, simplified for clarity}. Subsequently, models are trained to generate such sequences, and decoding them becomes straightforward. The key to the success of these models across various architectures lies in the incorporation of copy mechanisms. In the context of copy mechanism-based RE, the fundamental component is an LSTM that, at each time step, opts to select either a token from the source sequence or a limited additional vocabulary, such as punctuation or special tokens like \texttt{@gene}.

In more recent developments, the use of LLMs has emerged as a novel seq2seq approach for RE \cite{luo2022biogpt, gupta2025comparison}. The authors of BioGPT, for example, have fine-tuned their model using soft prompts to generate relations within natural language sentences, such as \texttt{The relation between <head entity> and <tail entity> is <relation type>}. These constructs with place holders for entities and relation types (also called predicates) are often called output \textit{templates}. The filled-in output template is then processed using regular expressions to extract the relations from the LM's generations. This method presents a significant advantage over traditional relation representation and copy-mechanism approaches primarily because it does not require mention annotations during training. Such a feature reduces the workload for annotators on additional datasets, as they can shift their focus solely to relation annotation rather than annotating every entity mention. Building on this, \citet{wadhwa-etal-2023-revisiting} modified this approach by designing target sequences as Python-interpretable tuples of relations, rather than in the form of natural sentences, for sentence-level RE tasks.

The remarkable performance of large, human-aligned language models in FS and ZS tasks has sparked interest in exploring their potential for low-resource RE. This emerging area of research particularly focuses on the capabilities of OpenAI's GPT models. \citet{wadhwa-etal-2023-revisiting} investigate the use of the instruction-finetuned GPT-3.5 for sentence-level biomedical RE. Their FS in-context learning experiments yield results that are competitive with  state-of-the-art approaches. In a similar vein, \citet{wang2023large} applied GPT-3.5 for sentence-level RC. Further advancing this line of inquiry, \citet{jahan-etal-2023-evaluation} conduct RE experiments using both GPT-3.5 and GPT-4, testing them on two RE dataset test sets, though in one they filter out all examples with no relations.

\subsection{Our contributions}

Given the effectiveness of ZS generation in other NLP tasks, in this paper, we investigate its utility in the high-value task of biomedical RE.  We comprehensively test the effectiveness of OpenAI GPT-4~\cite{openai2023gpt4}, OpenAI o1~\cite{jaech2024openai}, and the open-weights GPT-OSS~\cite{openai2025gptoss120bgptoss20bmodel} on seven RE datasets that vary in domain, length of text, diversity of entity and relation types, whether relations are entity-level (EL) or mention-level (ML), and whether relations are described across multiple sentences.   We analyze model errors to determine the strengths and weaknesses of this approach. The code, datasets, and the LLM prompts for all our experiments are available here: \url{https://github.com/bionlproc/ZeroShotRE}. 

Our research differs from prior studies in several ways. Previous research predominantly explored general biomedical tasks --- a valuable effort --- but restricting the study to one or two datasets is insufficient to explore the intricacies of RE. Our experiments encompass a broader spectrum of biomedical RE tasks, across seven datasets. We employ datasets that include relations confined within single sentences as well as those with relations spanning across multiple sentences. Moreover, some datasets we study feature relations between entities, while others contain relations between \emph{mentions} of entities. This variety introduces a range of complexities, including varying levels of difficulty and differing quantities of relations within each text. Such diversity underlines that, although these tasks are all categorized under RE, they each pose unique challenges to RE methodologies. Another crucial aspect of our work is the comprehensive evaluation of performance across entire test sets, facilitating direct comparisons with other studies, whether they are fine-tuned, FS, or ZS. %Next, we use features of the OpenAI API that restrict model output to be a valid JSON object, facilitating direct extraction of structured output.  Lastly, our study undertakes an in-depth analysis of the specific challenges posed by generative ZSRE.

\section{Materials and methods}
\vspace{-1mm}
\subsection{Task definition}
Let $\left(x,y\right)$ be an example in the test dataset, where $x$ is text and $y$ is the set of annotated relations expressed within $x$. Depending on the dataset, the text and relations may have different structures. $x$ is most commonly a title-abstract pair but maybe a sentence along with a larger passage containing it or a single string of text. $y$ depends on the dataset as well. In general, a relation consists of a set of typed entities along with a relation type connecting them. In most datasets, relations hold between two entities, but in the DCE \cite{tiktinsky-etal-2022-dataset} dataset, they hold between a variable number of entities. Entities and relations are typed, but the number of types varies by dataset. 

% Make ML vs EL figure?
Depending on the dataset, relations are either annotated at the mention level (ML) or entity level (EL). ML relations hold between textual mentions (exact spans) of entities. Textual mentions may consist of a span of text or multiple in the case of discontinuous entity mentions. EL relations hold between normalized entities, that is, the entities are provided in the form of an ID number corresponding to a biomedical concept from a controlled vocabulary. In most commonly used EL datasets (including all that we use), textual mentions of the biomedical concepts with their normalized ID number are also provided. Biomedical databases of relations are typically structured with EL relations. 

For ZSRE, we guide the LLM $\mathcal{M}$ to predict $y$ using template $T$. That is,

\begin{equation}
\widehat{y}=\mathcal{M}\left(T\left(x\right)\right),
\end{equation}
where $T$ is a user chosen natural language instruction along with an output template. 

\subsection{Extraction}
Most prior work on generative RE has relied on traditional supervision. In such scenarios, the choice of template is of moderate importance, because $\mathcal{M}$ is finetuned to learn the nature of the problem and the structure of the output. Without a supervision signal, it is challenging to guide $\mathcal{M}$ to (1) understand the nature of the problem and (2) output relations in a consistently structured form. The latter is important for biomedicine if RE is to be automated, and important for research as it permits performance metrics to be calculated. To address these challenges, $T$ adds a complete description of the RE task as well as instructions to produce output in the form of a JSON object, a description of the format the JSON object should take, and an example of what a filled-in JSON could look like.

Given that producing consistent structured output from a language model that has been principally trained to produce natural language is a problem faced in information extraction in general, a few attempts have been made to solve it \cite{clownfish2023, jsonformer} or produce a structured output posthoc \cite{kor}.  Given that we use GPT-4~\cite{openai2023gpt4}, o1~\cite{jaech2024openai} and GPT-OSS~\cite{openai2025gptoss120bgptoss20bmodel} as $\mathcal{M}$ and the fact that OpenAI  added functionality in their API for obtaining JSON objects as output in two different modes, we use their tools.  The first tool they developed requires the user to provide a schema (in the form of a JSON object) delineating the structure the output JSON should exhibit.  The more recent tool infers the schema from the prompt.  We refer to these modes as \emph{explicit} and \emph{inferred} modes and experiment with both on GPT-4. For OpenAI o1 and GPT-OSS \cite{jaech2024openai, openai2025gptoss120bgptoss20bmodel}, we test only the mode that performed better on GPT-4, due to budgetary constraints. Unlike GPT-4, the other two models are designed with test-time chain-of-thought based reasoning ability; it learns to ``recognize and correct its mistakes''  and `` break down tricky steps into simpler ones."\footnote{https://openai.com/index/learning-to-reason-with-llms/}
%The overall flow of our work is shown in Figure~\ref{fig: Zero shot Relation Extraction}. 

%\begin{figure*}
%\centering
%\includegraphics[scale=0.2]{images/workflow.png}
%\caption{\label{fig: Zero shot Relation Extraction} Overall flow of the zero-shot relation extraction using OpenAI API.}
%\end{figure*}

\subsection{Evaluation \label{subsec:evaluation}}

As a task, RE is complex, and there are many reasonable ways to measure performance.  This has led to a proliferation of measures, but also confusion and conflation of them --- so much so that rigorous study of the issue has been made \cite{taille2020let}. Unfortunately, the state of affairs has only worsened as (1) researchers have not heeded this work, (2) papers have faded descriptions of details necessary for reproducibility, (3) EL RE has been introduced, and (4) seq2seq methods, which have grown in popularity, lend themselves to new performance measures. 

Among the three main methods that have been published for EL RE --- JEREX \cite{eberts-ulges-2021-end}, seq2rel \cite{giorgi-etal-2022-sequence}, and BioGPT \cite{luo2022biogpt} --- no two calculate F1 in the same way.  JEREX measures performance very strictly: a predicted relation is considered correct if it matches a gold relation exactly, and entities within the relations are judged correct when mentioned boundaries are correct. Seq2rel's ``strict'' measure is similar, except that rather than entity mentions being judged on boundary correctness, they are judged on whether the predicted strings match gold entity mention strings, and duplicate gold mention strings are collapsed to a single mention.  JEREX correctness therefore implies seq2rel ``strict'' correctness, but not vice versa.  We note that since seq2rel uses a copy mechanism that points directly to tokens, nothing prevents them from making an exact comparison with JEREX.  However, \citet{giorgi-etal-2022-sequence} compared their performance with JEREX using slightly different ``strict'' metrics as indicated earlier.  They additionally use a ``relaxed'' measure of correctness that only requires a majority of predicted entity mentions to match that of a gold entity.  

The generative approach of BioGPT lends itself to the extraction of a single entity mention rather than all of them, and therefore  \citet{luo2022biogpt} deem a predicted relation correct if the extracted mentions match the \emph{longest} mention in the \emph{dataset}, rather than the example text.  Further distinguishing their performance measure from previous papers, \citet{luo2022biogpt} filter examples with no gold relations from the dataset.  Despite these differences, they compare their performance with seq2rel. At this point, it is not clear, on any dataset, which of these methods has the highest performance; nor is it clear that they all \emph{can} be compared with one another, even if done with utmost care.  To make matters worse, \citet{jahan-etal-2023-evaluation} do not describe their evaluation methodology or provide code.

\begin{table*}[htbp]
\centering
\small
\renewcommand{\arraystretch}{1.3}
\resizebox{\textwidth}{!}{%
\begin{tabular}{lccccc}
\toprule
\textbf{Datasets}  & \textbf{Type} & \textbf{Input Type} & \textbf{$\#$ Entity Types} & \textbf{$\#$ Predicates} & \textbf{Examples w/o relations}  \\
\midrule
ADE \cite{gurulingappa2012development}  & \multirow{4}{*}{Mention-Level} & Sentence & 2 & 1  & No     \\
DCE \cite{tiktinsky-etal-2022-dataset}   & & Abstract & 1 & 3  & Yes \\
ChemProt \cite{krallinger2017overview}  & & Abstract & 2 & 5  & Yes  \\
DDI \cite{herrero2013ddi}  & & Abstract & 4 & 4  & Yes   \\
\midrule
CDR \cite{li2016biocreative}  & \multirow{3}{*}{Entity-Level} & Abstract & 2 & 1  & No   \\
GDA \cite{wu2019renet}  & & Abstract & 2 & 1  & No  \\
BioRED \cite{luo2022biored}  & & Abstract & 4 & 8  & Yes   \\
\bottomrule
\end{tabular}
}
\caption{\label{table:datasets} Basic properties of the biomedical datasets tested, including whether relations were annotated at the mention-level or entity-level, whether the input text is a sentence or an abstract, the number of entity types and predicates, and whether the dataset contains instances with no relations.}
\end{table*}

Our method most closely resembles BioGPT, but we believe that an extracted entity mention matching any gold one should be considered correct; so we develop yet another performance measure and strive for the utmost clarity in explaining it.  In the EL RE context, we consider a predicted relation to match a gold relation if (1) each extracted entity mention participating in a relation matches any gold entity mention, (2) entity types are correct, and (3) relation type is correct (this is trivial when there is only one relation type.)  In the ML RE context, gold entities consist of a single mention, so (1) becomes simpler: an extracted entity mention must match the gold entity mention.  For EL RE datasets, we honor the annotation at the EL by mapping entities of predicted relations to their normalized ID numbers (based on gold annotations) and removing duplicate predictions before assessing matches to gold relations.  True positives are predicted relations matching gold relations; false positives are predicted relations that do not match any gold relations; and False negatives are unmatched gold relations. We calculate precision, recall, and F1-score for each dataset. 

\vspace{-1mm}
\subsection{Datasets \label{subsec:datasets}}

Table \ref{table:datasets} shows the basic properties of the datasets we studied.  Three of them contain EL relations; these datasets naturally contain relations with entity mentions across multiple sentences.  The remaining four datasets contain intra-sentence ML relations, though relation types may be more easily extracted when the surrounding context is available. 

The \textbf{\texttt{ADE}} dataset \cite{gurulingappa2012development} consists of sentences extracted from MEDLINE case reports describing adverse effects resulting from drug use, extracted from medical case reports. It contains two entity types: drugs and adverse effects and one relation type, adverse drug event. There is no official split of the dataset.

\textbf{\texttt{DCE}} \cite{tiktinsky-etal-2022-dataset} documents the efficacy of drug combination therapies, presenting a unique RE challenge in that relations contain a variable number of entity types.  Each instance consists of an abstract, within which a focal sentence is identified that contains multiple drug references. The drug references are classified as either being positive, for a beneficial drug combination, non-positive, for a combination with a neutral or negative effect, or non-combination, when the drugs are not given in combination.  Following the practice of the original authors, DCE performance is evaluated using two metrics: \texttt{Positive Combination F1} score and \texttt{Any Combination F1} score. The \texttt{Positive Combination F1} treats the relation type \texttt{positive} as the positive class, while the \texttt{Any Combination F1} score lumps \texttt{positive} and \texttt{non-positive} relation types together, and treats them as the positive class.

The primary aim of \textbf{\texttt{ChemProt}} \cite{krallinger2017overview} is to extract intra-sentence relations between chemical compounds and proteins/genes from biomedical abstracts. Relation types holding between these entities can be described as \texttt{upregulator}, \texttt{downregulator}, \texttt{agonist}, \texttt{antagonist}, or \texttt{substrate of}.  Over 25\% of abstracts contain no relations.

\textbf{\texttt{DDI}} \cite{herrero2013ddi} annotates intra-sentence interactions between four types of pharmacological substances: brand-name drugs, generic drugs, drug categories, and substances not approved for human use. Drug-drug interactions are either descriptions of pharmacokinetic mechanisms, descriptions of effect/pharmacodynamic mechanisms, recommendations about drug combinations, or documented interactions without additional details.  Nearly two-thirds of instances contain no relations.
The \textbf{\texttt{CDR}} \cite{li2016biocreative} and \textbf{\texttt{GDA}} \cite{wu2019renet} datasets respectively annotate diseases induced by chemicals/drugs or associated with genes in PubMed abstracts.  Both datasets contain EL relations with a single relation type.

\textbf{\texttt{BioRED}} \cite{luo2022biored} dataset annotates eight non-directional relation types holding between genes, gene variants, chemicals, and diseases. The relation types are \texttt{positive correlation}, \texttt{negative correlation}, \texttt{association}, \texttt{binding}, \texttt{co-treatment}, \texttt{drug interaction}, \texttt{comparison}, and \texttt{conversion}; certain relation types are only valid for a subset of all combinations of entity types. Instances in BioRED often contain many relations, sometimes in excess of 90.  Presumably, for this reason, there are only 100 test instances.

We show ZS prompts for \textbf{\texttt{ChemProt}} and \textbf{\texttt{CDR}} in Table \ref{table:prompts} in the Appendix.  Prompts for all seven biomedical datasets are made available here:  \url{https://github.com/bionlproc/ZeroShotRE/tree/main/prompts}.

\section{Results}

We present results for GPT-4, OpenAI o1, and GPT-OSS-120B in Table \ref{table:results}. Performance varies substantially across datasets, with consistently higher scores on ADE, DCE, CDR, and GDA compared to ChemProt, DDI, and BioRED. A clear pattern emerges: datasets with only 1 or 2 relation types yield much higher performance, while those with 4--8 relation types, often accompanied by a larger set of entity types, show lower performance. These observations may reflect spurious correlations; future work could test this hypothesis by slicing datasets by entity and relation types to examine whether performance improves.

When comparing models, GPT-4 generally underperforms relative to both o1 and GPT-OSS-120B. GPT-OSS consistently surpasses GPT-4 across all datasets, while o1 also outperforms GPT-4 except on ADE, CDR, and GDA—datasets that are simpler and contain only a single relation type. One explanation is that o1 and GPT-OSS-120B better accommodate tasks with greater relational and entity complexity through more robust reasoning. As a result, both surpass GPT-4 on ChemProt, BioRED, and DDI, each of which contains 4–8 distinct relation types. On average, o1 is over 3 F1 points better than GPT-4, with BioRED showing the largest relative gain (more than doubling GPT-4 performance). GPT-OSS-120B achieves the strongest overall results, attaining the best F1 scores on most datasets. It is encouraging to see a open-weights model that is better than large closed models (by an average four F1 points over the o1 model). Precision–recall trends are largely consistent across models, though o1 and GPT-OSS-120B tend to favor precision on ChemProt and DDI, while GPT-4 exhibits relatively higher recall.

\begin{table*}[ht!]
\centering
\renewcommand{\arraystretch}{1.5}
\resizebox{\textwidth}{!}{%
\begin{tabular}{l|*{12}{>{\centering\arraybackslash}p{1cm}}}
\toprule
\multirow{2}{*}{\textbf{Datasets}} & \multicolumn{3}{c}{\textbf{GPT-4 (Inferred)}} & \multicolumn{3}{c}{\textbf{GPT-4 (Explicit)}} & \multicolumn{3}{c}{\textbf{OpenAI o1 (Inferred)}} & \multicolumn{3}{c}{\textbf{GPT-OSS-120B (Inferred)}}\\
 &\textbf{$P$}& \textbf{$R$} &\textbf{$F_1$} &\textbf{$P$} &\textbf{$R$} &\textbf{$F_1$}  &\textbf{$P$} &\textbf{$R$} &\textbf{$F_1$} &\textbf{$P$} &\textbf{$R$} &\textbf{$F_1$}  \\ \midrule
ADE \cite{gurulingappa2012development} & 75.3 & 60.4 & 67.0 & 76.7 & 62.8 & 69.1  & 73.5 & 62.8 & 67.7 & 76.7 & 68.8 & \textbf{72.5}  \\
DCE (Pos.) \cite{tiktinsky-etal-2022-dataset} & 58.9  & 68.7   & 63.4  & 61.3 & 66.7  & 63.9   & 61.5 & 74.7  & 67.5 & 64.7 & 73.3 & \textbf{68.8}   \\
DCE (Any) \cite{tiktinsky-etal-2022-dataset} & 55.6 & 76.1 & 64.2 & 49.2 & 71.8 & 58.4 & 69.6 & 74.6 & 72.1 & 70.6 & 77.0 & \textbf{73.7} \\
ChemProt \cite{krallinger2017overview} & 24.1 & 24.6 & 24.3 & 19.7 & 23.0 & 21.2 & 37.0 & 20.7 & 26.5 & 28.9 & 31.0 & \textbf{30.0}\\
DDI \cite{herrero2013ddi} & 27.7 & 33.6 & 30.4 & 27.6 & 33.2 &  30.1 & 46.1 & 29.3 &  35.8 & 36.2 & 53.3 & \textbf{43.1} \\
\hline
CDR \cite{li2016biocreative} & 48.9 & 42.3 & 45.3 & 49.3 & 42.4 & {45.6}  & 46.6 & 41.2 & 43.7 & 52.1 & 48.5 & \textbf{50.2} \\
GDA \cite{wu2019renet} & 46.0 & 63.4 & 53.3 & 46.1 & 65.2 & {54.0}   & 40.2 & 57.6 & 47.3 & 45.5 & 67.4 & \textbf{54.4}  \\
BioRED \cite{luo2022biored} & 12.6 & 7.1 & 9.1 & 15.1 & 7.3 & 9.8  & 30.8 & 18.6 & \textbf{23.2} & 27.0 & 17.9 & 21.5  \\
\hline
\textbf{Average}  & 41.7 & 43.4 & 41.9 & 41.4 & 43.3 & 41.6  & 48.5 & 43.6 & {44.9} & 47.7 & 51.7 & \textbf{49.0}  \\
\bottomrule
\end{tabular}
}
\caption{\label{table:results} Main results for end-to-end ZSRE experiments.  As DCE is evaluated in two ways (see Section~\ref{subsec:datasets}), their performance values are averaged before being included in the calculation for the ``Average'' row. }%Since the GPT-4 Inferred and Explicit methods differ only slightly, we chose to apply only the Inferred method to OpenAI o1. In addition, we report results from GPT-OSS-120B (Inferred), which achieves competitive or superior performance in several datasets. Overall, OpenAI o1 and GPT-OSS-120B surpass GPT-4 in most cases.}
\end{table*}

As discussed in Section~\ref{subsec:evaluation}, a valid comparison of performance between current biomedical RE methods is generally not possible. However, all of the performance measures are obviously positively correlated, so we collate performance from other publications with reasonably transparent evaluation methodology in Table \ref{table:SOTA}. We find that \textit{supervised} methods using far smaller LMs perform similarly or better than our method, an unsurprising result \cite{wadhwa-etal-2023-revisiting}.  However, due to the aforementioned difficulties of comparison, the only obvious discrepancy occurs in ChemProt, where our method fared poorly. 
ChemProt encodes fine-grained and mechanistic relations between chemicals and proteins where the predicates are semantically close. This may require explicit learning of subtle lexical cues and biochemical context, which is difficult in the ZS setting, leading to the large performance gap relative to the supervised score.

\begin{table*}[htbp]
\centering
\small
\renewcommand{\arraystretch}{1.3}
\begin{tabular}{lccccccc}
\toprule
\textbf{Methods} & ADE & DCE  & ChemProt & DDI & CDR & GDA & BioRED \\
\midrule
Yan et al. \cite{yan2021partition} & 83.2 &  - & -  & - & - & - & -\\
Seq2Rel \cite{giorgi-etal-2022-sequence} & - & 66.7\textsuperscript{P$\dagger$}/71.1\textsuperscript{A$\dagger$} & - & - & 40.2\textsuperscript{$\dagger$}/52.4\textsuperscript{$\ddagger$}& 55.2\textsuperscript{$\dagger$}/70.5\textsuperscript{$\ddagger$} & - \\
BioGPT \cite{luo2022biogpt} & - & - & - & 40.8 & 46.2 & - & - \\
PURE \cite{zhong-chen-2021-frustratingly} &  - & - & 69.0  & - & - & - & -\\
\midrule
GPT-4 (zero-shot) & 69.1& 63.9\textsuperscript{P}/64.2\textsuperscript{A} & 24.3 & 30.4 & 45.6 & 54.0 & 9.8 \\
OpenAI o1 (zero-shot) & 67.7& 67.5\textsuperscript{P}/72.1\textsuperscript{A} & 26.5 & 35.8 & 43.7 & 47.3 & 23.2 \\
GPT-OSS-120B (zero-shot) & 72.5 & 68.8\textsuperscript{P}/73.7\textsuperscript{A} & 30.0 & 43.1 & 50.2 & 54.4 & 21.5 \\
\bottomrule
\end{tabular}
\caption{\label{table:SOTA} Comparison of performance (F1) of OpenAI ZS scores  with previous finetuned methods.  Note that F1 does not have identical meaning across methods (see Section~\ref{subsec:evaluation}). %We do not include any performance numbers from papers that do not describe the evaluation methodology clearly. 
Superscripts P and A refer to the ``Positive'' and ``Any Combination'' evaluation settings for DCE~\cite{jiang2023end}. Superscripts $\dagger$ and $\ddagger$ refer to the ``strict'' and ``relaxed'' evaluations described in Section~\ref{subsec:evaluation}.}
\end{table*}

We analyzed errors from all datasets to glean insights into the pitfalls of ZSRE.  We first note which aspects of RE were highly successful.  GPT-4 was nearly perfectly faithful to the structured schema we described in our templates and generated entity and relation types were nearly always selected from the set of types we described in the templates.  Predicted entity mentions were usually assigned the correct entity type as well.  Predicted mentions are rarely not present in the text. Last, it was uncommon for relation types to be incorrect when entities participating in relations were correct.

All models tend to under-predict relations when an instance contains more than a few gold relations. Figure~\ref{fig:many_golds} in the Appendix depicts this pattern for CDR on GPT-4, a representative example. It shows that the average number of relations predicted per test instance lags further behind the number of gold relations as the number of gold relations increases, and that this naturally results in decreased recall.  We attribute this to the observation that generative models tend to perform worse with long sequences \cite{hochreiter2001gradient, 10.1162/tacl_a_00531}.

% \begin{table*}[h!]
% \centering
% \renewcommand{\arraystretch}{1.3}
% \begin{tabular}{lccccccccc}
% \toprule
% \multirow{2}{*}{\textbf{Datasets}} & \multicolumn{3}{c}{\textbf{GPT-4 (Inferred)}} & \multicolumn{3}{c}{\textbf{OpenAI o1 (Inferred)}} \\
%  &\textbf{$P$}& \textbf{$R$} &\textbf{$F_1$} &\textbf{$P$} &\textbf{$R$} &\textbf{$F_1$}  \\ \midrule
% ADE \cite{gurulingappa2012development} & 75.3 & 60.4 & 67.0 & 73.5 & 62.8 & \textbf{67.7}  \\
% DCE (Positive combination) \cite{tiktinsky-etal-2022-dataset} & 58.9  & 68.7   & 63.4  & 61.5 & 74.7  & \textbf{67.5}   \\
% DCE (Any combination) \cite{tiktinsky-etal-2022-dataset} & 55.6 & 76.1 & 64.2 & 69.6 & 74.6 & \textbf{72.1} \\
% ChemProt \cite{krallinger2017overview} & 24.1 & 24.6 & {24.3} & 37.0 & 20.7 & \textbf{26.5} \\
% DDI \cite{herrero2013ddi} & 27.7 & 33.6 & 30.4 & 46.1 & 29.3 &  \textbf{35.8} \\
% \hline
% CDR \cite{li2016biocreative} & 48.9 & 42.3 & \textbf{45.3} & 46.6 & 41.2 & 43.7  \\
% GDA \cite{wu2019renet} & 46.0 & 63.4 & \textbf{53.3} & 40.2 & 57.6 & 47.3  \\
% BioRED \cite{luo2022biored} & 12.6 & 7.1 & 9.1 & 30.8 & 18.6 & \textbf{23.2}  \\
% \hline
% Average  & 41.7 & 43.4 & 41.9 & 48.5 & 43.6 & \textbf{44.9}  \\
% \bottomrule
% \end{tabular}
% \caption{\label{table:results-o1} Comparison of inferred scores between GPT-4 and OpenAI o1. Since the Inferred and Explicit methods differ only slightly, we chose to apply only the Inferred method to OpenAI o1. As a result, OpenAI o1 surpasses GPT-4 in most datasets.}
% \end{table*}

A common error across datasets and models we encountered was that of partial matching entity mentions, in which a predicted relation nearly matches a gold relation, but predicted mentions either include extra words not found in gold mentions or exclude words found in them.  For example, we extracted the incorrect chemical-disease relation  (\texttt{chemical: methamphetamine, disease: methamphetamine-induced psychosis}), which would have been correct had we extracted the disease as \texttt{psychosis}.  Future research should focus on extracting correct boundaries for entity mentions, as this was a major source of error. 

Frequently, false positives appear to be correct relations missed by the annotators.  This has been previously documented in RE datasets \cite{tran2019neural, tan2023revisiting} and has been shown to artificially  deflate performance. Given the high frequency of missed relations, it may be prudent to re-annotate biomedical benchmark RE datasets in the mode of \citet{tan2023revisiting}.

For the most part,  models predicted mentions that either were exact spans of source text or concatenated discontinuous spans.  However, in some cases, they used domain knowledge, predicting a text string not found in the source text.  In one extracted relation from GDA, we predicted the entity \texttt{interleukin-10}, which did not appear in the text in this form, whereas the gold version was \texttt{interleukin (IL)-10}.  

Besides error patterns holding across datasets, errors arose particular to specific datasets.  Our method largely failed on BioRED, with frequent, obviously incorrect, predicted entity mentions and types as well as relations claiming opposite relationships between two given entities.  In DCE, drug combinations were frequently missing drugs that participate in a relation.  Also, relation types were often incorrectly assigned; we suspect that this is caused by the domain-specific knowledge, like the interpretation of lab result quantities, sometimes required to correctly assign relation~type.

\section{Discussion}
\label{sec-disc}
In the LLM era, a fundamental question is whether and in what settings can we simply use ZS predictions from frontier LLMs without the tedious and expensive creation of training data and custom supervised models. The answer to this is heavily dependent on particular task on hand in terms of expectations on recall and precision and the consequences of false positives/negatives. For example, if the relations are being used for knowledge discovery, focusing more on precision can minimize creation of misleading hypotheses. However, to conduct systematic reviews on information encoded in the relations (e.g., drug--side-effects), the relations extracted should have high recall. 

As a first step to assess the general end-to-end ZSRE competence of LLMs, we created a new benchmark and conducted experiments with frontier LLMs. Our high level takeaway is that for shorter instances with fewer relations (less information density), ZSRE is closer to fully supervised models; this is more so if the entities are shorter (fewer tokens) such as mostly single token drug entities in DCE and ADE datasets. Entity complexity is also high in ChemProt dataset, which may have contributed to the vast discrepancy (40 points in F1) between supervised and ZS performances. Relatively, high density longer inputs (e.g., BioRED) lead to almost unusable performance at this point. Further assessments that account for partial matches of entities may be warranted but unless that is done carefully, the results may not be meaningful. For instance, partial matches that do not involve the head word of an entity phrase are mostly incorrect and misleading at best. %And hence simple partial matching that is typically employed where even a single token match is considered correct is not helpful. 

All results using OpenAI GPT models for publicly available datasets since GPT-3~\cite{brown2020language} come with a caveat: we do not know what data the models were trained on.  These LLMs train on massive amounts of scraped web data, and most datasets we used are available on the web in some form.  It is possible that they may have been trained on these datasets. However the very low scores obtained for BioRED and ChemProt indicate that this contamination is unlikely. Please note that here we are not focusing on the textual inputs (sans labels) being part of the LLM pre-training corpora; this was shown to not cause any contamination in general~\cite{li2024open}. The focus here is on whether particular task-specific labels were part of the training process. Recently, methods to detect membership of specific texts in the pretraining corpora of LLMs have been introduced~\cite{rastogi2025stamp} but foolproof tools that assess whether ground truth labels of supervised tasks are leaked are not available. Recent evidence shows that label contamination is possible for question answering tasks where the answer is a single word or a short phrase, but if rich, structured outputs (as in end-to-end RE) are required, memorization is not as prevalent~\cite{wang2025generalization}. 
While label contamination should be kept in mind, that potential should not be grounds for not evaluating ZS performances; other teams have been exploring the same with OpenAI models~\cite{li2023revisiting,zhang2024study}, albeit not in an end-to-end manner. In light of recent calls for action~\cite{jacovi2023stop}, we have encrypted the datasets using in this project before hosting them in our GitHub space. Others can still reproduce our results by running our scripts, which would decrypt them on the fly in a programmatic manner.   

\section{Limitations}
Our results and associated implications apply only to English datasets, but we hope others will follow up with benchmarks in other languages. With regards to LLMs studied, we only considered OpenAI models while there are more frontier options (e.g., from Anthropic, Google, and xAI), which we could not work with due to time and cost constraints. Also, there are datasets with more complex annotation schemes than BioRED such as the cancer genetics and pathway curation datasets of the BioNLP 2013 shared task~\cite{nedellec2013overview}. However, the download links for them are not active and additional efforts are needed to carefully recover and experiment with them. Finally, as we  already acknowledged in Section~\ref{sec-disc}, there is some potential risk of label contamination although we believe it is minimal for RE tasks where memorization is nontrivial.  

\section*{Acknowledgment}
This work is supported by the U.S.~NIH National Library of Medicine through grant R01LM013240.  The content is solely the responsibility
of the authors and does not necessarily represent the official views of the NIH.

\bibliography{sample}
\appendix
\section{Appendix}
\label{sec:appendix}
\begin{figure*}
\begin{center}
\includegraphics[scale = 0.6]{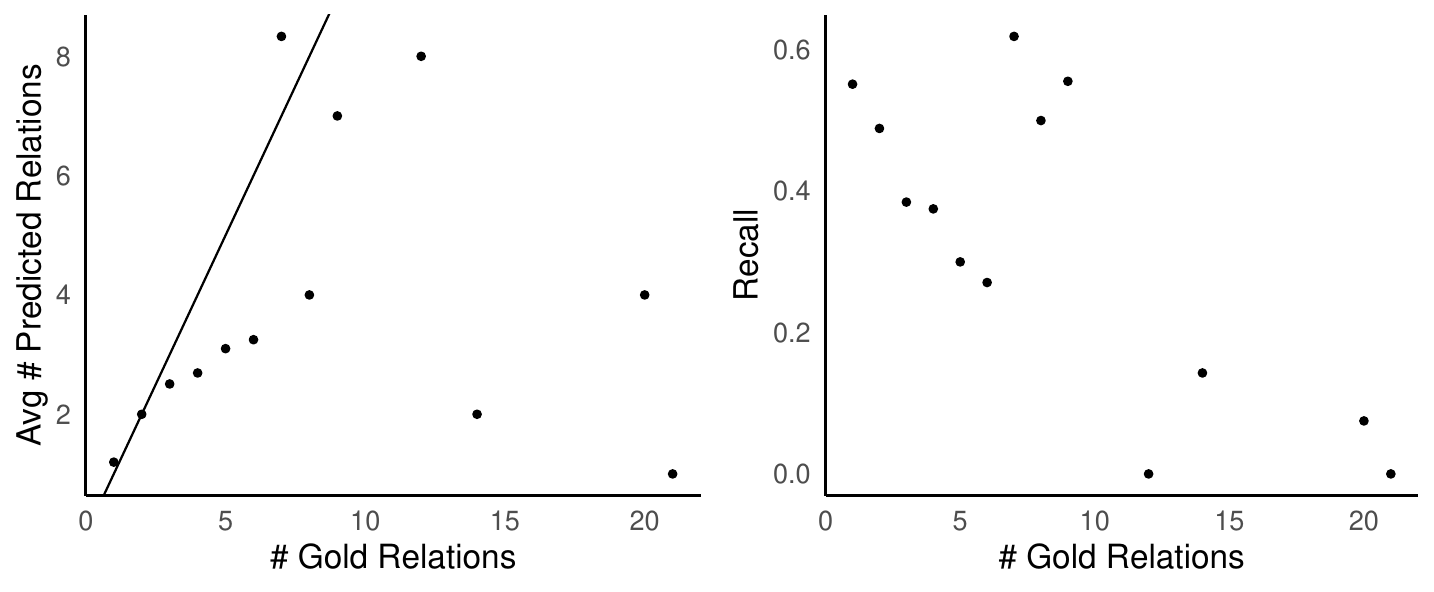}
\caption{\label{fig:many_golds} (Left) The average number of GPT-4 predicted relations per test instance is plotted against the number of gold relation in the instance for the CDR dataset.  The line $y = x$ is overlayed for ease of interpretation.  (Right) Recall is calculated for subsets of the data by the number of gold relations.}
\end{center}
\end{figure*}

In Table~\ref{table:prompts}, we give examples of ZS prompts for \textbf{\texttt{ChemProt}} and \textbf{\texttt{CDR}} datasets each describing the task, predicate label definitions, along with JSON output templates with some dummy filled in examples.
In Figure~\ref{fig:many_golds}, we show how the number of predicted relations does not keep up as the number of gold relations increases (x-axis) and hence recall decreases as the number of gold relations increase in an instance.
\begin{table*}[ht!]
\centering
\tiny
\setlength{\tabcolsep}{1.75pt} 
\begin{tabular}{p{1cm}p{14cm}}
\toprule
\textbf{Dataset} & \textbf{Prompt} \\
\midrule
ChemProt  & Your task is to extract all relevant triples from an input biomedical text. Each triple has a chemical mention, a gene/protein mention, and a predicate linking the two mentions. The predicate belongs to one of the following 5 predicates: ``CPR:3'', ``CPR:4'', ``CPR:5'', ``CPR:6'' and ``CPR:9''. These 5 predicates are further specified as below: \newline
``CPR:3'' includes UPREGULATOR, ACTIVATOR and INDIRECT UPREGULATOR \newline
``CPR:4'' includes DOWNREGULATOR, INHIBITOR and INDIRECT DOWNREGULATOR \newline
``CPR:5'' includes AGONIST, AGONIST ACTIVATOR and AGONIST INHIBITOR \newline
``CPR:6'' includes ANTAGONIST \newline
``CPR:9'' includes SUBSTRATE, PRODUCT OF and SUBSTRATE PRODUCT OF \newline
Note that chemical or gene/protein mentions should have appeared from the original input text. Make sure that each relation is based on mentions within the same sentence in an abstract. \newline
The output triples should be saved as per the following format: \newline
\{``relations'': \newline
[ \newline
  \{``chemical'': ``chemical1'', \newline
    ``gene'': ``gene1'', \newline
    ``relation'': ``relation1''\}, \newline
  \{``chemical'': ``chemical2'', \newline
    ``gene'': ``gene2'', \newline
    ``relation'': ``relation2''\}, \newline
  ... \newline
] \newline
\} \newline
The output will be \{``relations'':[]\} if there are no relevant triples expressed in the input text. \newline
With this format, a hypothetical example output for a biomedical text could be the following: \newline
\{``relations'': \newline
[ \newline
  \{``chemical'': ``polyamines'', \newline
    ``gene'': ``caspase'', \newline
    ``relation'': ``CPR:3''\}, \newline
  \{``chemical'': ``DL-alpha-difluoromethylornithine'', \newline
    ``gene'': ``ornithine decarboxylase'', \newline
    ``relation'': ``CPR:4''\}, \newline
  \{``chemical'': ``putrescine'', \newline
    ``gene'': ``ODC'', \newline
    ``relation'': ``CPR:9''\} \newline
] \newline
\}
\\ 
\midrule
CDR  & Your task is to extract all chemical-disease relations from a text in which the chemical/drug induces the disease. Note that the chemical or disease names should have appeared in the original input text. \newline
The output should be saved as per the following format: \newline
\{``relations'': \newline
[ \newline
  \{``chemical'': ``chemical1'', \newline
    ``disease'': ``disease1''\}, \newline
  \{``chemical'': ``chemical2'', \newline
    ``disease'': ``disease2''\}, \newline
  ... \newline
] \newline
\} \newline
The output will be \{``relations'':[]\} if there are no chemical-disease pairs in which the chemical induces the disease expressed in the input text. \newline
With this format, a hypothetical example output for a biomedical text could be the following: \newline
\{``relations'': \newline
[ \newline
  \{``chemical'': ``Lidocaine'', \newline
    ``disease'': ``cardiac asystole''\}, \newline
  \{``chemical'': ``daunorubicin'', \newline
    ``disease'': ``neutropenia''\} \newline
] \newline
\}
\\
\bottomrule
\end{tabular}
\caption{\small{ZS prompts for \textbf{\texttt{ChemProt}} and \textbf{\texttt{CDR}}. All prompts for seven biomedical datasets are released in our GitHub website.}}
\label{table:prompts}
\end{table*}

\end{document}